\documentclass{article}

\usepackage{appendix}

\usepackage[final]{neurips_2025}
\makeatletter
\renewcommand{\@notice}{}
\makeatother

\usepackage[utf8]{inputenc} 
\usepackage[T1]{fontenc}    
\usepackage{hyperref}       
\usepackage{url}            
\usepackage{booktabs}       
\usepackage{amsfonts}       
\usepackage{nicefrac}       
\usepackage{microtype}      
\usepackage{xcolor}         
\usepackage{enumitem}
\usepackage{caption}
\usepackage{amsmath}
\usepackage{graphicx}
\usepackage{placeins}
\usepackage{fancyhdr}

\title{PanoWorld-X: Generating Explorable Panoramic Worlds via Sphere-Aware Video Diffusion}

\author{
\begin{tabular}{c}
Yuyang Yin\textsuperscript{1}\footnotemark[1],
HaoXiang Guo\textsuperscript{2}\footnotemark[1], 
Fangfu Liu\textsuperscript{3},
Mengyu Wang\textsuperscript{1},
Hanwen Liang\textsuperscript{4},\\
Eric Li\textsuperscript{2}, 
Yikai Wang\textsuperscript{5}\footnotemark[2],
Xiaojie Jin\textsuperscript{1},
Yao Zhao\textsuperscript{1}, 
Yunchao Wei\textsuperscript{1}\footnotemark[2]
\end{tabular} \\[0.5em]   
\normalfont
\textsuperscript{1} Beijing Jiaotong University \quad
\textsuperscript{2} Skywork AI \quad
\textsuperscript{3} Tsinghua University\\
\textsuperscript{4} University of Toronto \quad
\textsuperscript{5} Beijing Normal University
}

\begin{document}
\maketitle
\footnotetext[1]{Equal contribution.}
\footnotetext[2]{Corresponding authors.}

\begin{figure}[h]
    \centering
    \begin{minipage}{0.8\textwidth}
        \centering
        \captionsetup{type=figure}
        \vspace{-5mm}
        \includegraphics[width=1\linewidth]
        {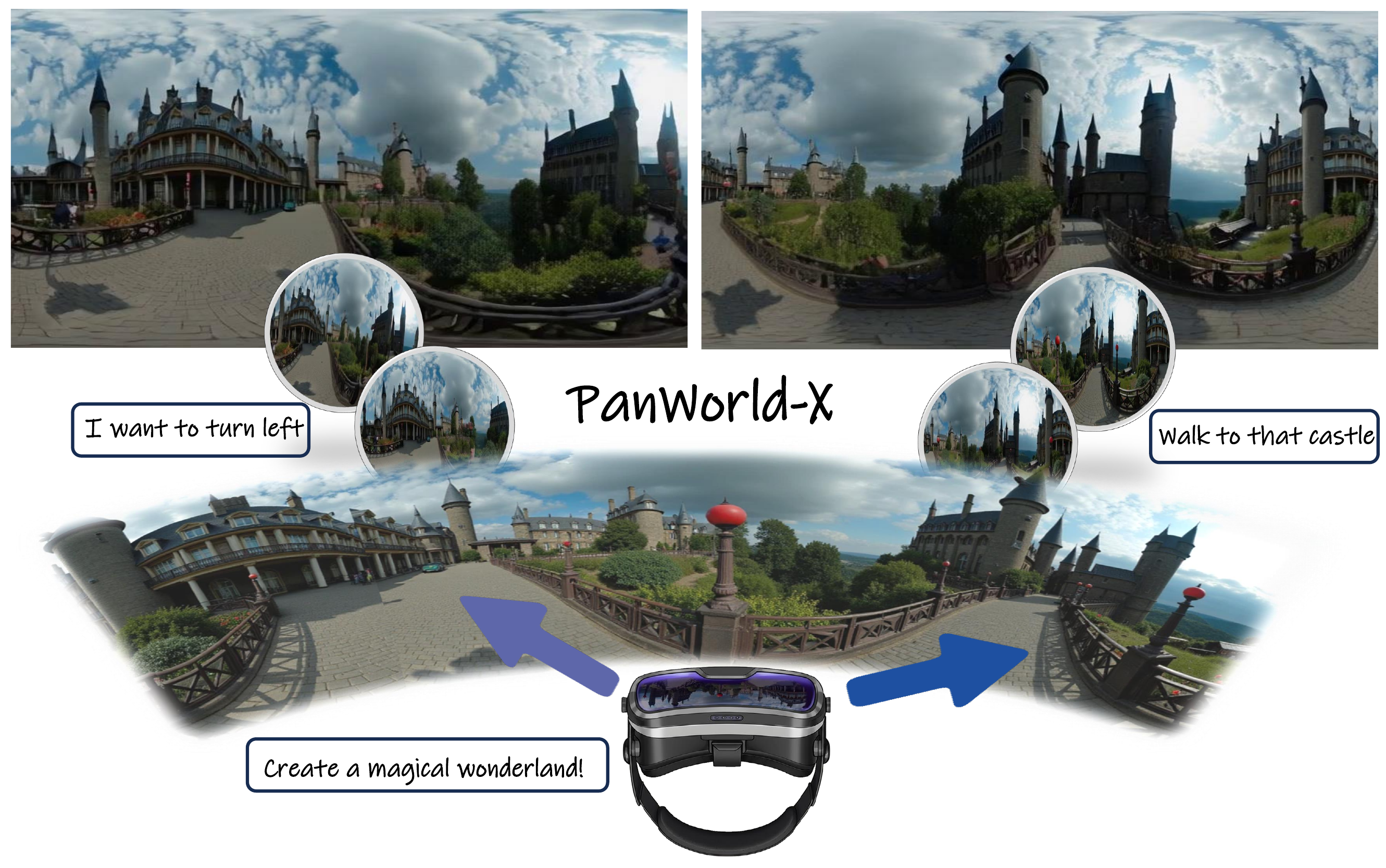}
        \vspace{-2mm}
        \caption{\textit{PanoWorld-X} is a novel framework for high-fidelity and controllable panoramic video generation with diverse exploration route. }
        \label{fig:teasr}
    \end{minipage}
\end{figure}

\begin{abstract}
  Generating a complete and explorable 360-degree visual world enables a wide range of downstream applications. While prior works have advanced the field, they remain constrained by either narrow field-of-view limitations, which hinder the synthesis of continuous and holistic scenes, or insufficient camera controllability that restricts free exploration by users or autonomous agents. To address this, we propose \textit{PanoWorld-X}, a novel framework for high-fidelity and controllable panoramic video generation with diverse camera trajectories. Specifically, we first construct a large-scale dataset of panoramic video-exploration route pairs by simulating camera trajectories in virtual 3D environments via Unreal Engine. As the spherical geometry of panoramic data misaligns with the inductive priors from conventional video diffusion, we then introduce a Sphere-Aware Diffusion Transformer architecture that reprojects equirectangular features onto the spherical surface to model geometric adjacency in latent space, significantly enhancing visual fidelity and spatiotemporal continuity. Extensive experiments demonstrate that our PanoWorld-X achieves superior performance in various aspects, including motion range, control precision, and visual quality, underscoring its potential for real-world applications. Project page: \textcolor{blue}{\linebreak\url{https://yuyangyin.github.io/PanoWorld-X/}}.

\end{abstract}
\section{Introduction}
\label{sec:intro}

The physical world is a 360-degree, fully explorable spatial environment where observers can perceive their surroundings from any angle. As an observer moves, the visual scene dynamically adapts to their changing position.
Replicating these characteristics in the digital domain is critical to support diverse applications, such as the creation of immersive virtual reality (VR) environments~\cite{yang2024dreamspace,xanthidou2024collaboration} for human users and the development of simulated training spaces~\cite{Sora} for embodied intelligence~\cite{liu2024aligning,ma2024survey} and autonomous driving agents~\cite{mao2023language,wang2024omnidrive}.
Thus, the generation of wide-field-of-view, explorable virtual worlds has emerged as a prominent research focus.

Towards this challenging objective, a technically feasible approach is to train large-scale video generation models~\cite{yang2024cogvideox,kong2024hunyuanvideo,wang2025wan} using extensive real-world data. Taking Sora~\cite{Sora} as a prime example, it is considered a promising technology for simulating real-world environments. Nevertheless, these models usually can only generate perspective videos with a limited field of view, which restricts their capability to fully capture and represent entire scenes. Although certain methods attempt to generate long videos~\cite{kim2024fifo,song2025history} or 3D-aware videos~\cite{liu2024reconx,sun2024dimensionx,yu2024viewcrafter} with technical strategies to include more scene information, their outcomes still fall short in terms of geometric consistency and scene scale.
Consequently, some researchers have shifted their focus to panoramic content generation. Thanks to its inherent characteristics, a 360-degree scene can be efficiently represented by a single image via simple equirectangular projection (ERP).


Previous studies on panorama generation have achieved significant progress, mostly focusing on static panoramic images~\cite{zhang2024taming,li2023panogen,wu2023panodiffusion,yang2024layerpano3d,feng2023diffusion360,ye2024diffpano}. However, static imagery fails to address environmental incompleteness issues caused by occlusion.
Panoramic video generation remains underexplored, with only a few studies~\cite{wang2024360dvd,tan2024imagine360,li20244k4dgen,liu2024dynamicscaler} tackling this challenge, yet several limitations persist:
Firstly, \textbf{limited scenario diversity and restricted camera motion} in these works lead to poor performance in rendering occluded elements or distant small objects.
Secondly, \textbf{lack of precise control over panoramic content} (e.g., specifying movement trajectories) limits interactivity with users or agents.
Thirdly, \textbf{inadequate generation quality, characterized by spatiotemporal incoherence}, hinders generalization to new scenarios and real-world applications.
We further identify that these limitations primarily arise from three key challenges:
\textbf{(1) Scarcity of high-quality data}: Currently, only the WEB360 dataset~\cite{wang2024360dvd} is publicly available, but it suffers from limited scale, scene diversity, and video motion dynamics.
\textbf{(2) Coarse-grained task definitions}: Although studies like~\cite{ye2024diffpano} attempt to render training data with diverse camera trajectories, they still rely solely on textual input for generation, lacking finer-grained control mechanisms.
\textbf{(3) Neglect of panoramic geometry}: Existing methods treat panoramic data as perspective data, directly applying pre-trained perspective diffusion models without accounting for inherent geometric characteristics (e.g., spherical pixel distribution).
To address these challenges, we introduce \textbf{PanoWorld-X}, a novel framework for generating high-quality, explorable panoramic videos under movement signals control. Our approach starts by constructing a large-scale and content-diverse dataset of panoramic videos paired with exploration routes, built via 3D scene rendering in Unreal Engine (UE). We design an exploration route sampling strategy to automatically sample camera viewpoint trajectories in 3D scenes, filter invalid paths using collision detection, and render panoramic videos along valid trajectories. We further employ the Video-LLaMA3~\cite{zhang2025videollama} to filter low-quality samples and generate textual captions.
As a result, we obtained 116,759 high-quality panoramic videos, each paired with its corresponding 3D exploration route, which we refer to as the \textbf{PanoExplorer dataset}.

Subsequently, we introduce Explorable Sphere-Aware DiT Block to replace the original DiT block of pre-trained video diffusion models, in order to achieve explorable and high-quality panoramic video generation.
This block consists of three attention branches: the original global attention from the pre-trained diffusion model, an Exploration-Aware Attention introducing exploration route signal control, and a Sphere-Aware Attention to enhance spherical geometric perception and improve generation quality.
The original global attention branch remains frozen, which enhances the efficiency and stability of training.
Exploration-Aware Attention processes viewpoint trajectories by first encoding them into pixel-wise Plücker embeddings~\cite{sitzmann2021light}, which are then fused with image tokens to enhance the association between trajectories and visual content. Trajectory information is then refined through attention layers and integrated into the main branch through a zero-initialized projection layer. 

Sphere-Aware Attention addresses the prior misalignment between panoramic and perspective data. Specifically, in perspective images or videos, adjacent regions usually contain related information, prompting the pre-trained models to prioritize nearby tokens. This enhances local semantic and texture modeling but causes issues when applied to panoramic content.
For example, the left/right edges and polar regions in the ERP panoramas are physically connected in 3D space but spatially separated in 2D projection images, causing pre-trained models to underweight their strong semantic correlations and degrade coherence.
To resolve this, we reproject ERP panoramas onto a spherical surface and compute spatiotemporal spherical distances between patches. Sphere-Aware Attention then acts as a parallel branch that only attends to tokens within a spherical distance threshold, thus enhancing the neglected implicit correlations on the sphere.

We conduct comprehensive experiments to evaluate the proposed framework. The results demonstrate that, compared to existing panoramic video generation models and camera controllable generation models, our PanoWorld-X exhibits superior performance in various aspects, including generation quality, motion range, and control precision. In summary, our key contribution includes:
\begin{itemize}[itemsep=0em, topsep=0em, leftmargin=1em]
    \item We collect a large number of panoramic videos and proposed a refined data collection and processing pipeline, resulting in a high-quality, large-scale, and content-diverse panoramic video dataset paired with exploration routes.
    \item We introduce an Explorable Sphere-Aware DiT Block, which is composed of the newly proposed Exploration-Aware Attention and Sphere-Aware Attention to achieve trajectory controlment and improve the spatiotemporal coherence of generative results.
    \item Experimental results demonstrate that our approach outperforms previous methods and achieve large-scale movement, precise controllability, and high-quality panoramic video generation.
\end{itemize}

\section{Related Work}
\textbf{Video Diffusion Models.} Image generation models~\cite{ho2020denoising,rombach2022high} have achieved remarkable results and enabled numerous downstream applications~\cite{yin2023cle,yin20234dgen,huang2024classdiffusion,wang2025characonsist,wang2024region,wang2023segrefiner,ren2025videoworld,lin2025aligngen,hu2025dcedit,hu2024diffusion,jiao2024clip}.
Early research on video diffusion models \cite{blattmann2023stable,guo2023animatediff,chen2023videocrafter1} primarily focused on extending pre-trained image diffusion models \cite{rombach2022high} to video generation. These studies achieved dimensional extension by incorporating temporal interactions into the UNet architecture of image diffusion models. However, the limited scale of both models and datasets constrained the quality of the generated results.
More recently, Diffusion Transformers (DiTs) \cite{peebles2023scalable} were introduced, with authors demonstrating that Transformer-based \cite{vaswani2017attention,dosovitskiy2020image} generators exhibit superior scalability compared to UNet-based architectures. Over the past year, several DiT-based text-to-video models \cite{yang2024cogvideox,kong2024hunyuanvideo,wang2025wan} have emerged. Their successful scaling has significantly improved the performance in various aspects, including generated duration, motion amplitude, and temporal consistency, thus providing a more robust prior for a wide range of downstream tasks.

\textbf{View-Controllable Video Generation.}
Generating video that matches a sequence of changed views is a critical step toward creating a virtual world and applying it to scenarios such as agent systems. Existing research can be broadly categorized into two main types.
The first one \cite{wang2024motionctrl,he2024cameractrl,liang2024wonderland,bahmani2024ac3d,liu2024reconx,sun2024dimensionx,yu2024viewcrafter,li2025martian,liang2024diffusion4d} focuses on incorporating camera extrinsic parameters as additional input signal into pretrained video generation models, allowing the generated video to have perspective changes according to the position and orientation of the camera. The second type \cite{valevski2024diffusion,che2024gamegen} aims to generate video games, achieving video-game-like interactivity. The viewpoint changes in these works are controlled by simulating inputs from devices such as keyboards.
However, these methods face significant limitations due to the constrained information provided by perspective-view videos. They struggle to generate complete scenes while maintaining 3D consistency and cyclic consistency, leading to notable instability in the generated scenes.

\textbf{Panorama Generation Models.}
Benefiting from the rapid advancements in 2D image generation, diffusion-based panorama image generation models~\cite{zhang2024taming,li2023panogen,wu2023panodiffusion,yang2024layerpano3d,feng2023diffusion360,ye2024diffpano} has achieved significant results. Despite the 360-degree visible nature of panoramas, physically occluded content (e.g., what lies around the next corner) remains difficult to obtain. Consequently, Panoramic video is necessary to capture broader spatial information, facilitating the construction of a comprehensive world model. 360DVD~\cite{wang2024360dvd} initially established the WEB360 panoramic video dataset and accomplished text-to-Panoramic Video generation. Methods such as 4K4DGen~\cite{li20244k4dgen}, TiP4Gen~\cite{xing2025tip4gen},  Imagine360~\cite{tan2024imagine360} and others~\cite{liu2024dynamicscaler,tan2024imagine360,dong2025panolora,yang2025matrix} achieved dynamic panoramic video generation with object movements. However, these approaches are unable to generate content with significant viewpoint progression, restricting the scope of world content generation to relatively limited areas. 
Furthermore, these methods fail to achieve precise and controllable 360-degree worlds, which hinders the ability to interact effectively with the generated worlds. 
The most related work is GenEX~\cite{lu2024genex}. However, it does not incorporate the geometric properties of panoramas and instead simply fine-tunes the model, resulting in insufficient quality in the final details.
Therefore, our objective is to generate explorable panoramic videos that incorporate extensive spatial movement. 

\begin{figure}
    \begin{center}\vspace{-0.2cm}
        \includegraphics[width=0.98\linewidth]
        {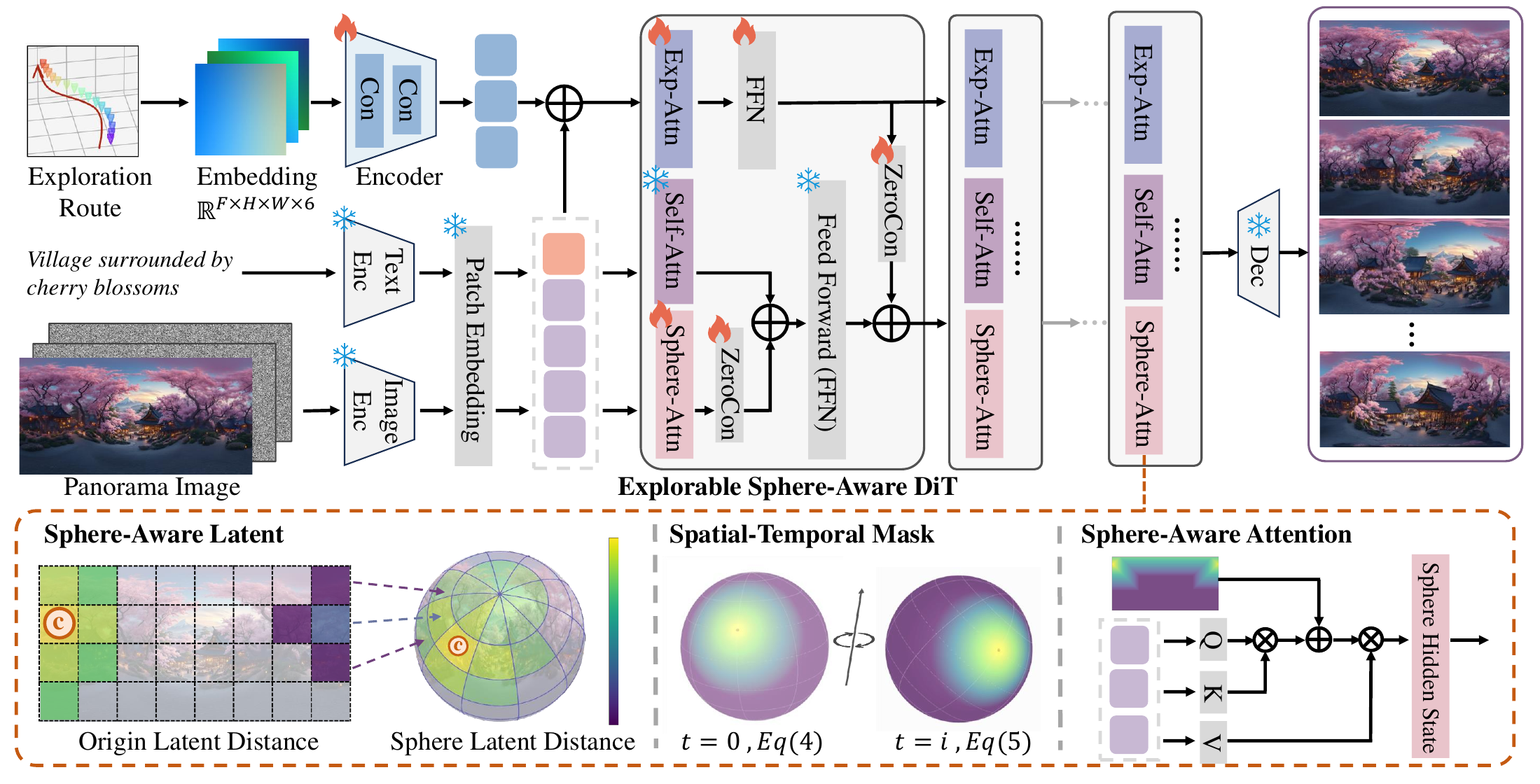}
        \vspace{-1mm}
        \caption{\textbf{PanoWorld-X Framework.} Given a panorama image with an exploration route, our model achieves high-quality and controllable panoramic video generation through the novel Explorable Sphere-Aware DiT blocks. This block employs the Exploration-Aware Attention to achieve precise view control, and the Sphere-Aware Attention to enhance spatiotemporal coherence by finding the related tokens through spherical geometry. For example, the upper-left token and the upper-right token are distant in the original latent space, but they are actually connected in reality. Therefore, we compute the distance on the sphere and use it as the basis for calculating the attention mask, which improves the interaction between associated tokens.}
        \label{fig:method}
    \end{center}
    \vspace{-0.3cm}
\end{figure}

\section{Method}
\label{sec:method}
To achieve high-quality panoramic video generation with free exploration, we first curate the PanoExplorer dataset in Section~\ref{sec:Data Curation} and propose an Explorable Sphere-Aware DiT block in Section~\ref{sec:control diffusion}, which serves as a replacement for the original DiT block in pre-trained models.

\subsection{Preliminary}
\label{sec:preliminary}
\textbf{Video Diffusion Model.}
The latest text-to-video generation models adopt latent-space multimodal DiT architectures, which consist of three key components: text encoders \cite{radford2021learning,raffel2020exploring} for encoding textual prompts, a 3D-VAE \cite{yu2023language,yang2024cogvideox} for video data compressing and tokenization, and a Transformer-based generator.
As the core component, the generator is composed of numerous transformer blocks. It takes flattened video tokens and text tokens as inputs, modeling video and textual information through 3D self-attention~\cite{yang2024cogvideox,kong2024hunyuanvideo}. Specifically, the input tokens are first transformed into query $Q$, key $K$, and value $V$ using linear blocks. The attention output is  computed as
\begin{equation}
Attn(Q,K,V)=softmax(\frac{\mathbf{Q} \mathbf{K}^T}{\sqrt{d}} )*\mathbf{V},
\end{equation}
where $d$ represents the dimension of the attention heads, used for normalizing the attention scores. Such architecture brings significant flexibility, allowing tokens from different spatiotemporal locations within a video to directly interact via attention.

However, this architecture presents certain challenges. On one hand, it incurs high computational and memory costs, making it more difficult to introduce additional control signals to guide the generation process. On the other hand, despite its global receptive field, the model sometimes overlooks essential long-range dependencies, resulting in decreased geometric consistency in the generated outputs.
For panoramic video, which exhibits substantial distribution differences compared to common visual data, these issues become even more pronounced.


\subsection{Data Curation}
\label{sec:Data Curation}
Collecting static panoramic videos with real-world exploration trajectories is challenging due to high labor costs and uncontrollable environments. To address this, we introduce the \textbf{PanoExplorer Dataset}, a scalable synthetic dataset for explorable Panoramic Video generation. Built entirely in Unreal Engine, it ensures high-fidelity simulation, diverse environments, and precise trajectory control, with five key construction stages.

\textit{Step 1: Data Collection in Unreal Engine.}
Unreal Engine enables flexible data preparation across diverse scenes, perspectives, and content. Using 504 high-fidelity 3D scenes, we cover varied indoor and outdoor environments with different weather and lighting, forming the basis for trajectory-based video recording.

\textit{Step 2: Exploration Route Sampling.}
We design a trajectory sampling algorithm to create plausible, visually coherent camera paths. For each scene, we first extract walkable surfaces (e.g., roads, floors) and apply Delaunay triangulation algorithm, which constructs a set of non-overlapping triangular meshes from sparse points on a two-dimensional plane. We then sample path candidates through three steps: (1) randomly selecting two mesh points, (2) computing the shortest path with Dijkstra's algorithm, and (3) applying Laplacian smoothing to reduce abrupt turns. Only trajectories over 18 meters are retained to ensure sufficient temporal dynamics.

\textit{Step 3: Collision Detection.}
We introduce a collision detection mechanism to eliminate trajectories causing ``geometry clipping'' or ``object intersections'', which degrade quality and stability. We employ a bounding box proxy algorithm, in which objects are simplified to 3D bounding boxes based on their nearest and farthest points, balancing spatial accuracy and computational efficiency. Trajectories are simulated step-by-step, and any intersecting paths are discarded.

\textit{Step 4: Spatial Normalization.}
To handle scene variations in size and depth, we extract the depth information directly from Unreal Engine and fix the physical distance between adjacent frames at 10 cm. This ensures sensitivity to spatial scale across small indoor and expansive outdoor environments. Each route is defined as:
$Er_o^T = (x_0^T, y_0^T, z_0^T, \alpha_0^T, \beta_0^T, \gamma_0^T)$,
where $x_i$, $y_i$, $z_i$ are spatial coordinates, and $\alpha_i$, $\beta_i$, $\gamma_i$ are yaw, pitch, and roll angles. Ablation studies confirm this normalization improves control fidelity and structural coherence.

\textit{Step 5: Data Annotation and Quality Filtering.}
We ensure dataset quality through two filtering stages:
(1) Automatic Filtering: We employ Video-LLaMA3~\cite{zhang2025videollama} to assess videos based on detailed quality, semantic information, and motion richness, and subsequently filter out low-quality content.
(2) Manual Assessment: The first frame of each video is manually screened to eliminate samples with poor rendering quality or missing details.
Finally, Video-LLaMA3 automatically annotates videos to support text-controlled and multimodal tasks.


After this multi-step pipeline, we retain 116,759 high-quality static panoramic video sequences, each paired with its corresponding 3D exploration route. We hope that our data curation pipeline and scalable dataset pipeline will facilitate future research.

\subsection{Explorable Sphere-Aware DiT Block}
\label{sec:control diffusion}
In this section, we propose the Explorable Sphere-Aware DiT Block, replacing the original DiT block in pre-trained video diffusion models. This block effectively integrates viewpoint variation control signals of the exploration process with enhanced perception of spherical geometric features in panoramic data, enabling the generation of high-quality, explorable panoramic videos.

\textbf{Exploration Route Representation.}
Most movement commands can be uniformly represented as a six-degree-of-freedom control signal, expressed as $Er_o^T=(x_0^T, y_0^T, z_0^ T,\alpha_0^T, \beta_0^T, \gamma_0^T)$. To achieve precise controllable generation, we aim to transform this control signal into a pixel-level positional representation. We select Plücker embeddings~\cite{sitzmann2021light} as the representations. The position information can be represented as a translation vector $tr_0^T$ originating from the origin of the world coordinate space, while the Euler angles can be transformed into a rotation matrix $R_0^T$. Additionally, the intrinsic matrix $K$ can be derived from the render camera information. The Plücker embeddings of the pixel $(u, v)$ of time $t$  can be expressed as $(tr \times d_{u,v}, d_{u,v})$, where $d_{u,v}$ denotes the direction vector calculated as $d_{u,v} = R K^{-1} [u, v, 1]^{T} + tr$. In general, the exploration route can be represented as an embedding matrix $Er \in \mathbb{R}^{T \times H \times W \times 6}$, which captures spatiotemporal motion information.

\textbf{Exploration-Aware Controllable Branch.}
Pretrained diffusion transformer(DiT) models are trained on vast video datasets, making full fine-tuning computationally expensive. 
Therefore, we aim to design an exploration route controllable branch that fulfills two critical goals: 1) enabling fine-tuning of large-scale diffusion transformer models with minimal data, and 2) maintaining the original latent space of the diffusion transformer to ensure output quality.

To achieve these objectives, we follow the principles of ControlNet~\cite{zhang2023adding}. We first encode the exploration route embedding using several 3D convolution layers to compress features into the same shape as the DiT latent.
Then, we concatenate the video latent and condition latent to enable comprehensive information interaction. We initialize a new Exploration-Aware Attention (Exp-Attn) module with the same parameters as the original DiT block. To enhance training efficiency and minimize the impact on the original model, we employ a zero-linear layer, ensuring that the exploration route controllable branch initially has no effect on the original branch. Finally, we perform an element-wise addition of the outputs from the controllable branch and the original branch to achieve deep integration of information from both branches.

\textbf{Spherical Geometric Representation.}
Previous studies~\cite{wang2024360dvd,lu2024genex} fine-tune perspective generative models~\cite{blattmann2023stable,rombach2022high} on panoramas, treating it as a process of adapting models to a specialized data domain.
However, this straightforward fine-tuning strategy neglects the inherent properties of panorama data, as the pixels in panoramas are actually distributed in a sphere, which has significantly different geometric characteristics compared to plane geometry.
Therefore, we introduce the following spatiotemporal distance representation based on spherical geometry, which has a crucial impact on the subsequent attention process.

In existing generative model frameworks, images or videos are encoded into a series of tokens. Previous 3D self-attention mechanisms enable the capture of spatially proximate tokens to gather more contextual information. Typically, different positional distances are determined based on planar geometry, where the distance between two points $p_1 = (x_1, y_1)$ and $p_2 = (x_2, y_2)$ is measured using the Euclidean distance:
\begin{equation}
d_{\text{Euclidean}}(p_1, p_2) = \sqrt{(x_2 - x_1)^2 + (y_2 - y_1)^2}. \label{equ:dist}
\end{equation}

While in panoramic ERP images, pixels distributed on a spherical surface introduce significantly different properties.
For example, the point on the leftmost in Fig.~\ref{fig:method} is physically connected with the points on the rightmost and those near the polar region in the spherical coordinates.
However, when unfolded into a flattened image, these areas exhibit large spatial coordinate distances as in Eq.~(\ref{equ:dist}).
Therefore, we introduce the spherical distance (Great-circle Distance) to accurately measure the distance between two points on the sphere.
For an ERP image of width $W$ and height $H$, a pixel at coordinates $(x, y)$ can be converted to spherical coordinates $(\theta, \phi)$ as follows
\begin{equation}
\theta = \frac{2\pi x}{W} - \pi, \quad \phi = \frac{\pi y}{H} - \frac{\pi}{2}.
\end{equation}
Here, $\theta$ represents the longitude (ranging from $-\pi$ to $\pi$), and $\phi$ represents the latitude (ranging from $-\frac{\pi}{2}$ to $\frac{\pi}{2}$).
Building on geometric priors, we reproject the ERP panorama image onto a spherical surface and recalculate the distances. For two points with spherical coordinates $(\theta_1, \phi_1)$ and $(\theta_2, \phi_2)$, the spherical distance $d_{\text{spherical}}$ is given by the Haversine formula:
\begin{equation}
d_{\text{spherical}}(p_1, p_2) = 2R \cdot \arcsin\left(\sqrt{\sin^2\left(\frac{\phi_2 - \phi_1}{2}\right) + \cos(\phi_1) \cos(\phi_2) 
\sin^2\left(\frac{\theta_2 - \theta_1}{2}\right)}\right).
\label{eq:haversine}
\end{equation}

From a temporal perspective, movement in the real world can be interpreted as the rotation of the panoramic sphere in various directions. We define a rotation matrix $\mathbf{R}^{t_i}(\alpha, \beta, \gamma)$ that rotates a point on the sphere by Euler angles $\alpha$ (yaw), $\beta$ (pitch), and $\gamma$ (roll) at time $t_i$ relative to its initial orientation at time $t_0$. For a point $\mathbf{p}^{t_i} = (\theta, \phi)$ on the sphere, the rotated point $\hat{\mathbf{p}}^{t_i}$ relative to $t_0$ is given by

\begin{equation}
\hat{\mathbf{p}}^{t_i} = \mathbf{R}^{t_i} (\alpha, \beta, \gamma) \cdot \mathbf{p}= \mathbf{R}_z^{t_i} (\alpha) \cdot \mathbf{R}_y^{t_i} (\beta) \cdot \mathbf{R}_x^{t_i} (\gamma)\mathbf{p},
\end{equation}
where $\mathbf{R}_z(\alpha)$, $\mathbf{R}_y(\beta)$, and $\mathbf{R}_x(\gamma)$ are the rotation matrices around the z-axis, y-axis, and x-axis, respectively.So we measure the point $\mathbf{p}_1^{t_0}$ and $\mathbf{p}_2^{t_i}$ distance by first rotate to $\hat{\mathbf{p}}_2^{t_i}$ then calculate the sphere distance given in Eq.~\eqref{eq:haversine}.


\textbf{Sphere-Aware Attention.}
Based on the above analysis, we introduce Sphere-Aware Attention (Sphere-Attn) to enhance the generative model's perception of spherical geometric features. Specifically, we redefine the distances between latent representations based on their positions on the spherical surface and attempt to enhance the response of closely located latent blocks.
We set the attention mask to 1 for regions where the distance is below a specific threshold $\tau$, indicating areas that require mutual reinforcement. Mathematically, the spatial-temporal attention mask $M$ is defined as
\begin{equation}
M(p_1^{t_i}, p_2^{t_j}) = \begin{cases}
1 & \text{if } d_{\text{spherical}}(p_1^{t_i}, p_2^{t_j}) \leq \tau, \\
0 & \text{otherwise}. \label{equ:mask}
\end{cases}
\end{equation}

Then we replicate the components of the original 3D self-attention and convert video tokens to query $Q$, key $K$, and value $V$. Given the sphere-aware attention mask, our sphere-aware attention can be computed as
\begin{equation}
SphereAttn(Q,K,V,\mathbf{M})=softmax(\frac{\mathbf{Q} \mathbf{K}^T}{\sqrt{d}} + \mathbf{M})*\mathbf{V}.
\end{equation}


As illustrated in Fig.~\ref{fig:method}, we employ a parallel mechanism. The video path simultaneously feeds into both the original self-attention block and the newly trained sphere-aware attention block. To ensure the sphere-aware attention block does not affect the initial training outcomes, a zero-initialized linear block is utilized. Compared to a sequential design, this parallel structure avoids significantly disrupting the output of each layer, which would otherwise lead to reduced training efficiency and increased difficulty in convergence.

\section{Experimentation}
\subsection{Implementation Detail}
\label{sec:Implementation Detail}
Since our model is based on a video diffusion framework, we select the advanced diffusion transformer model, CogVideoX-5B-I2V~\cite{yang2024cogvideox}. We fine-tuned the model to generate 49 frames with a resolution of $480 \times 720$. Due to the panoramic nature, the width-to-height ratio must adhere to 1:2. Therefore, we resized the output frames to $480 \times 960$. Compared to training a model with a native 1:2 aspect ratio, this post-processing resizing approach maximally preserves the prior knowledge of the original model. 
The model is trained on 8 A100 GPUs for 8000 iterations with the controllable branch and an additional 2000 iterations specifically for the sphere-aware attention block. During the inference stage, we first generate a panorama image using a text prompt with FLUX~\cite{flux2024}, leveraging the panorama LoRA released by Yang~\cite{yang2024layerpano3d}. Subsequently, we input the first image along with specific action signals into our model.

\subsection{Evaluation Datasets and Metrics}
\label{sec:metrics}
To evaluate the performance of explorable panoramic video generation, we randomly select 200 panoramic videos from our curated dataset. Each panoramic video includes a movement trajectory. We compare the generated videos with ground-truth video clips using multiple metrics:(1) Pixel-level visual quality is measured using PSNR, SSIM, and LPIPS. (2) Visual quality and temporal coherence are assessed using Frechet Inception Distance (FID)~\cite{heusel2017gans} and Frechet Video Distance (FVD)~\cite{unterthiner2019fvd}. (3) Exploration route control precision is evaluated using Rotation Error ($R_{err}$) and Translation Error ($T_{err}$) metrics, as introduced by \cite{he2024cameractrl}. These metrics compute the camera extrinsic parameters in comparison to the ground truth camera pose. To accommodate varying output lengths, the mean of these metrics is calculated rather than their sum.

\begin{figure}[t]
    \begin{center}
        \includegraphics[width=0.85\linewidth]
        {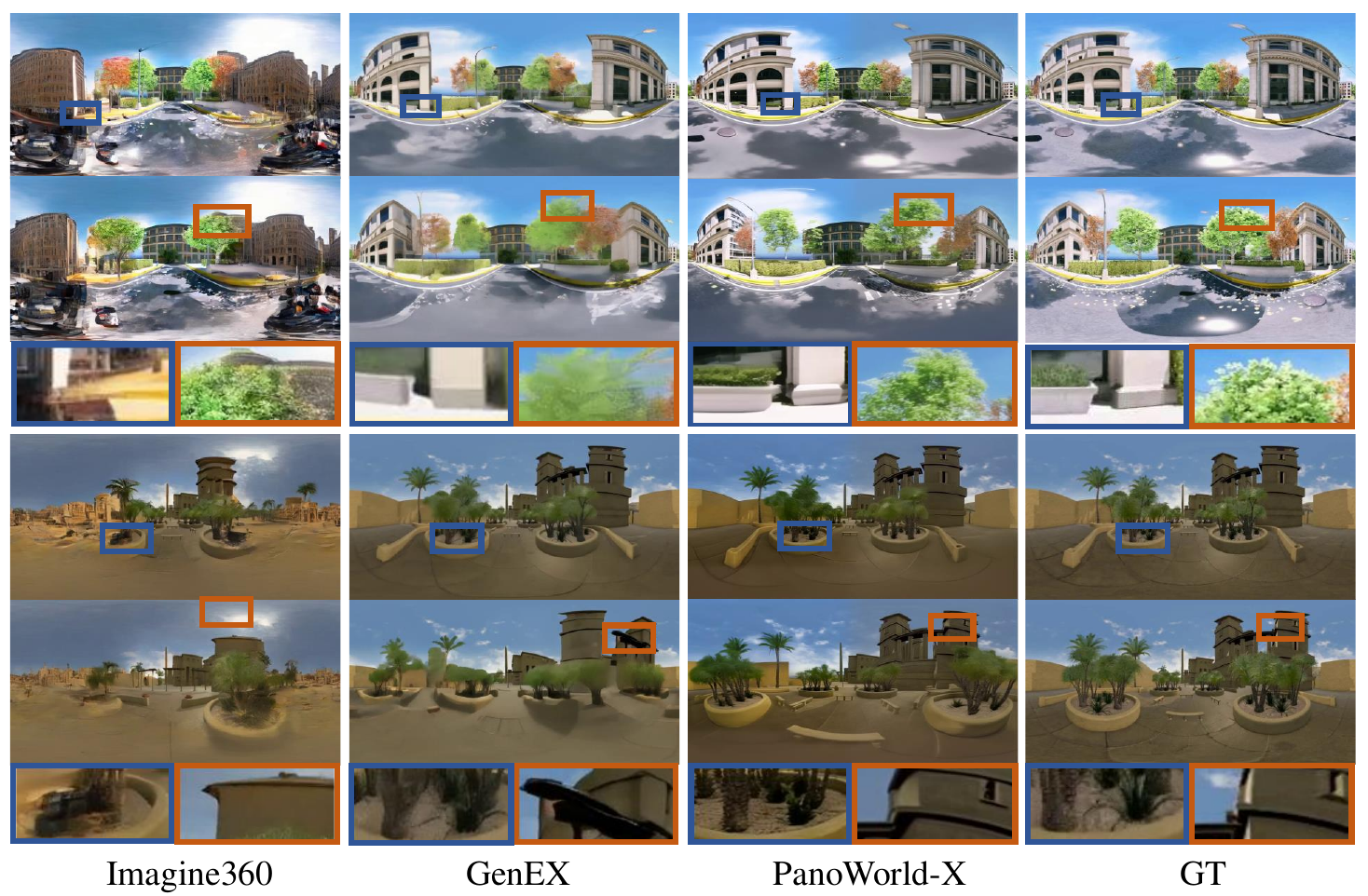}
        \vspace{-2mm}
        \caption{Qualitative comparisons with panoramic video generation models on generated keyframes. Our results exhibit superior detail clarity and geometric coherence throughout the movement.}
        \label{fig:pan}
    \end{center}
    \vspace{-2mm}
\end{figure}
\begin{figure}[t]
    \begin{center}
        \includegraphics[width=0.98\linewidth]
        {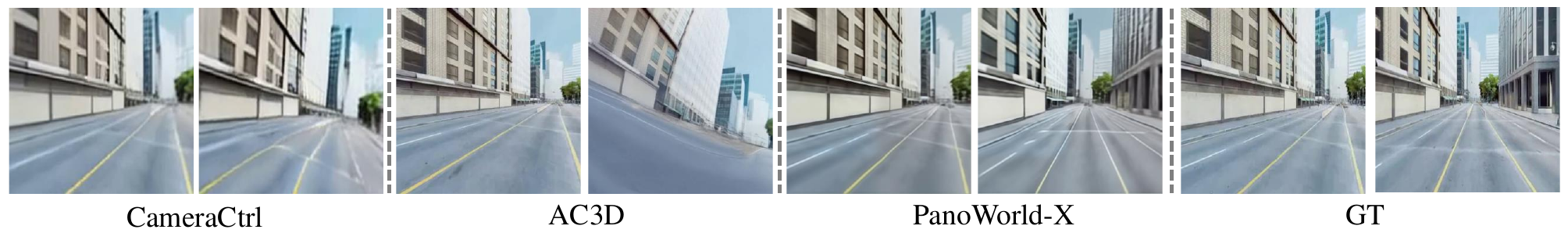}
        \vspace{-2mm}
        \caption{Qualitative comparisons with camera controllable methods on cropped perspective views from panoramas. Our results demonstrate more precise camera control and generation quality.}
        \label{fig:cam}
    \end{center}
    \vspace{-4mm}
\end{figure}

\subsection{Quantitative Results}
\textbf{Comparison with Panoramic Video Generation Models.}
We evaluate the performance of our proposed methods by comparing them with state-of-the-art Panoramic Video generation models. Specifically, we select three baseline methods for comparison:(1) 360DVD~\cite{wang2024360dvd} uses a trainable 360-Adapter to extend standard T2V~\cite{guo2023animatediff} models to the panorama domain. (2) Imagine360~\cite{tan2024imagine360} is a model designed to convert perspective videos into panoramic videos. We crop perspective videos from the ground truth videos and evaluate their generation capabilities. (3) GenEX~\cite{lu2024genex} is a stable video diffusion ~\cite{blattmann2023stable} based model for Panoramic Video generation. Based on its official checkpoint, it only supports image-to-video generation and lacks a controllable video generation checkpoint. Therefore, we only compare its video quality with our proposed methods.

The qualitative comparison shown in Fig.~\ref{fig:pan} demonstrates the high-quality generation capability of our model. Imagine360 struggles to consistently generate stable content on both sides, often leading to detail degradation. GenEX tends to produce blurry images during the generation process. In contrast, our method not only maintains excellent detail in the generated images but also achieves a significantly wider range of camera movement compared to other approaches. Additionally, Tab.~\ref{tab:combined} confirms that our method outperforms previous methods across all evaluation metrics.








\textbf{Comparison with Camera Controllable Generation Models.}
To evaluate our explorable route-controllable Panoramic Video generation, there are no existing works with the same setting for direct comparison. As a result, we compare our approach with camera-controllable generation models. Since these models are not trained on panoramic datasets, we crop our results into perspective view videos and input the ground truth perspective view images into camera-controllable models to ensure a fair comparison.
We select two state-of-the-art methods for this evaluation: CameraCtrl~\cite{he2024cameractrl} and AC3D~\cite{bahmani2024ac3d}.
We evaluate the generation results from two perspectives. First, in terms of image quality, as shown in Tab.~\ref{tab:combined}, our method surpasses previous approaches. The quality of details is also evident in the comparison provided in Fig.~\ref{fig:cam}. Second, regarding controllability, CameraCtrl often struggles to achieve precise control over content, typically allowing only very limited movement. While AC3D shows improvement in controllability compared to earlier methods, it still performs poorly when handling complex trajectories. In contrast, our method demonstrates a significant enhancement in controllability compared to previous approaches.

\begin{figure}[h]
    \vspace{-1mm}
    \includegraphics[width=0.98\linewidth]{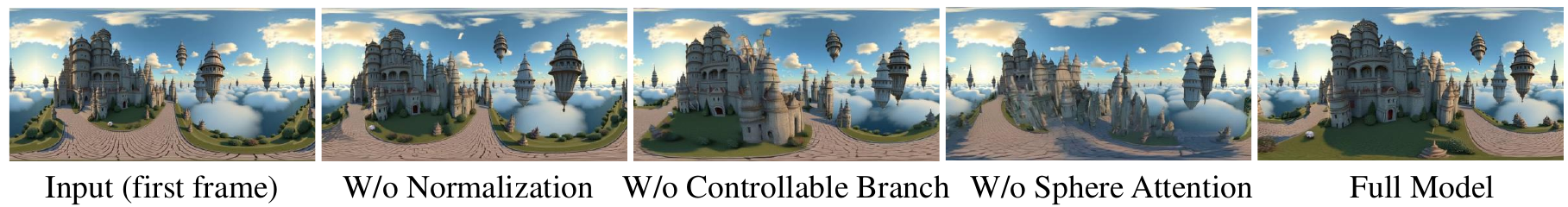}
    \vspace{-2mm}
    \caption{Ablation study of different components of our proposed framework.}
    \label{fig:abl}
\end{figure}

\textbf{Ablation Study.}
We primarily investigate the importance of the proposed components from three aspects.
 \textbf{(1) Data Position Normalization:} As mentioned in Sec~\ref{sec:Data Curation}, we normalize the dataset stride scale, which enhances sensitivity to control signals and facilitates more significant camera movements.
 \textbf{(2) Exploration Route Controllable Branch:} 
Previous works were unable to precisely control the direction of generated content movement. With this module, we can achieve forward movement in the generated panorama video while enabling rotation in any direction. Quantitatively, this approach significantly reduces both $R_{err}$ and $T_{err}$.  \textbf{(3) Sphere-Aware Attention:} To fully utilize the geometric properties of panorama data, the sphere-aware attention mechanism references the most relevant patches on the spherical surface. Experiments demonstrate that this module effectively preserves the geometric integrity of the panoramic content during video generation. From a quantitative perspective, the results demonstrate significant improvements across all metrics.
This improvement is visually demonstrated in Fig.~\ref{fig:abl}. In summary, our PanoWorld-X not only enables content generation with free direction and magnitude but also effectively maintains fine-grained quality in panoramic videos.

\begin{table}[t]
    \centering
    \caption{Comparison of Panoramic Video Generation  and Camera Controllable Generation Models.}
    \label{tab:combined}
    \setlength{\tabcolsep}{5pt}
    \begin{tabular}{lcccccccc}
        \toprule
        {Models} & {PSNR} $\uparrow$ & {SSIM} $\uparrow$ & {LPIPS} $\downarrow$ & {FID} $\downarrow$ & {FVD} $\downarrow$ & {$R_{err}$} $\downarrow$  & {$T_{err}$} $\downarrow$ \\
        \midrule
        \multicolumn{8}{l}{\textit{Panoramic Video Generation Models}} \\
        \midrule
        360DVD~\cite{wang2024360dvd}          &         10.66            &   0.35                  &          0.73              &    111.85                   &            2049.21        &    --                 &         --           \\
        Imagine360~\cite{tan2024imagine360}      &    11.62                &  0.39                 &       0.591               &     66.73                 &  1830.46                  &    --             \    & --                    \\
        GenEX~\cite{lu2024genex}           &   16.12                 &  0.59               &   0.42                      &    42.22                   &    1113.72              &              --      &    --                 \\
        PanoWorld-X      &    \textbf{19.34}                &  \textbf{0.63 }               &  \textbf{  0.24 }                 &  \textbf{ 28.01 }                  &   \textbf{  467.18  }         &          --           &    --                 \\
        \midrule
        \multicolumn{8}{l}{\textit{Camera Controllable Generation Models}} \\
        \midrule
        CameraCtrl~\cite{he2024cameractrl}          &   11.56                  &   0.38                  &           0.61               &       108.12              &         2017.95           &        0.097           &     0.245                \\
        AC3D~\cite{bahmani2024ac3d}      &       13.77              & 0.49                  &  0.52                     &     41.98                 &          842.29          &              0.081       &         0.087            \\

        PanoWorld-X Perspetive      &  \textbf{16.76}                  &             \textbf{ 0.56}     &       \textbf{  0.42}             &          \textbf{   38.63 }       &        \textbf{ 586.51}       &        \textbf{  0.061}           &       \textbf{ 0.073}             \\
        \bottomrule
    \end{tabular}
    \vspace{-3mm}
\end{table}

\begin{table}[t]
    \centering
    \caption{Ablation Study on Individual Components.}
    \label{tab:ablation}
    \setlength{\tabcolsep}{4pt}
    \begin{tabular}{lccccccc}
        \toprule
        {Models} & {PSNR} $\uparrow$ & {SSIM} $\uparrow$ & {LPIPS} $\downarrow$ & {FID} $\downarrow$ & {FVD} $\downarrow$ &{$R_{err}$} $\downarrow$  & {$T_{err}$} $\downarrow$ \\
        \midrule
        w/o Position Normalization      &   17.11                 &         0.55          &         0.32             &     40.37               &      751.18       &0.114   & 0.102      \\
        w/o Controllable Branch          &        16.30             &       0.53              &      0.36                  &   38.71                    &    769.42&           0.102 &0.152     \\

        w/o Sphere-Aware Attention           &   17.59                 &       0.56          &      0.27               &         29.96          &    492.98          & 0.069 & 0.076          \\
        Full model     &     \textbf{19.34}               &   \textbf{0.63}               &           \textbf{   0.24  }      &  \textbf{28.01}&   \textbf{467.18  }       &\textbf{0.061 }        & \textbf{ 0.073   }           \\
        \bottomrule
    \end{tabular}
    \vspace{-5mm}
\end{table}

\section{Conclusion}
\label{sec:Conclusion}
We introduce PanoWorld-X, a novel framework designed for generating explorable panoramic videos. It addresses two key limitations of prior methods: the narrow field of view in traditional video generation models and the issues of uncontrollable camera movements and limited motion range in existing panorama generation approaches. Leveraging a curated dataset, we design a controllable branch to enable precise exploration route control and employ a sphere-aware attention mechanism to enhance visual quality. Our evaluations demonstrate that PanoWorld-X outperforms previous state-of-the-art methods. We believe our work will inspire future research in this domain.

\textbf{Limitation.}
Although our method demonstrates the capability to generate explorable panoramic video worlds and achieves favorable results compared to other approaches, several limitations remain. First, the current model architecture does not support the generation of long video content. Second, our framework currently can only perceive exploration routes as input, while incorporating more interactive features in the future work is important to improve the user experience.

\bibliographystyle{plain}
\bibliography{neurips_2025}




\end{document}